# Object-Oriented Knowledge Extraction using Universal Exploiters

Dmytro Terletskyi
Faculty of Computer Science and Cybernetics
Taras Shevchenko National University of Kyiv
Kyiv, Ukraine
dmytro.terletskyi@gmail.com

*Abstract* — **This paper contains analysis and extension of exploiters-based knowledge extraction methods, which allow generation of new knowledge, based on the basic ones. The main achievement of the paper is useful features of some universal exploiters proof, which allow extending set of basic classes and set of basic relations by finite set of new classes of objects and relations among them, which allow creating of complete lattice. Proposed approach gives an opportunity to compute quantity of new classes, which can be generated using it, and quantity of different types, which each of obtained classes describes; constructing of defined hierarchy of classes with determined subsumption relation; avoidance of some problems of inheritance and more efficient restoring of basic knowledge within the database.**

## I. INTRODUCTION

During recent years application of knowledge-based systems has extremely increased, therefore variety of systems and knowledge bases for different domains were developed. In spite of this, the invention of efficient methods for knowledge representation (KR), inference and extraction is still topical issue.

Nowadays there are many knowledge representation formalisms (KRFs), which are used for knowledge-based systems (KBSs) development. Currently the most commonly-used approaches are semantic networks, ontologies, logical and rule-based formalisms. However, the certain programming paradigm, language and some stack of programming technologies should be chosen for development of a KBS. This choice is very important, because each programming paradigm and language provides certain tools for system development and determined mechanisms of interaction among its modules, in particular interaction with database. Thus, chosen KRF and programming technologies for its implementation, should be at least compatible with respect to each other. Otherwise, developed KBS will have complicated interaction between the level of KRF and the level of its implementation. Consequently it can decrease the efficiency of such system. Despite this, chosen formalism should provide efficient representation of hierarchically-structured knowledge about particular domain, because concepts hierarchy makes KR more compact and allows performing of reasoning over itself. Furthermore, the hierarchy should be stored in the database in such way, that KBS can be able to extract the knowledge efficiently and represent them in terms of programming language, using which the system was developed. However, the representation of hierarchies is possible, only if chosen KRF and programming language support mechanism of inheritance.

Currently, the most commonly used programming paradigm is an object-oriented programming (OOP). All OOP-languages and many KRFs support single inheritance. However, as it was shown in [1]–[3], inheritance mechanism causes problem of exceptions, redundancy and ambiguity, which usually arise during construction of hierarchies and reasoning over them.

## II. KNOWLEDGE EXTRACTION

According to [4]–[7], knowledge extraction is defined as creation or acquiring of knowledge from structured (e.g. relational databases, object-oriented database models, UML, XML and their fuzzy extensions, proposed in [8], [9]), semi-structured (e.g. infoboxes) and unstructured (e.g. text, documents, images) data sources. In addition, the extracted knowledge should be represented in machine-processable format that enables inference.

According to [6], there are two main paradigms of KE: ontology-based and open domain. They also can be called as close world knowledge extraction and open world knowledge extraction. The idea of first approach is to use ontology as vocabulary, which defines the types of concepts used in the knowledge base. It means that knowledge base contains defined number of types of entities and relationships. Thus, only relations included in the vocabulary can be extracted from the knowledge base.

In the second approach, knowledge-based system does not have any vocabulary and pre-specified relationship types in the knowledge base. It means that each entity or relation in knowledge base can be considered as a candidate. Therefore, any possible relation or assertion in the knowledge base can be extracted.

However, Unbehauen, Hellmann, Auer, Stadler et al. in [4], [5] argued about absence of clear definition of what extracted knowledge is and paid attention to the fact, that mere usage such KRFs as RDF/OWL can not sufficiently define the notion of «knowledge». They have formulated two important questions:

1. *What is the result of data representation in terms of RDF/OWL (triplification process)? Structured data or represented knowledge?*
2. *When does structured data became knowledge?*

Analyzing these questions, it is possible to conclude that result of such knowledge extraction, first of all, will be structured data, which then can be interpreted as some knowledge. However, such interpretation can be performed only using particular KRF, where notion of knowledge is defined in a proper way. Therefore, any KRF can be considered as interpreter of data, according to its own specifics and specifics of particular domain, for representation of which the formalism was developed.

One of attempts to solve earlier mentioned problems is such KRF as object-oriented dynamic networks (OODN), which was proposed in [10]. It provides representation of knowledge in OOP-like style and is compatible with respect to many OOP-languages. In addition, as it was demonstrated in [3], OODN allow constructing of polyhierarchies and avoiding, in many cases, problems of inheritance, which were mentioned above. Moreover, OODN have fuzzy extension, proposed in [11], [12], which provides representation of vague and imprecise knowledge, using the same structure as for the crisp case. One more feature of OODN is exploiters-based knowledge extraction (KE) methods, which provide generating of finitely defined set of new classes of objects and finitely set of new relations among them, based on the set of basic classes and relations among them. It allows calculation of quantity of new classes, which can be extracted, and quantity of different types, which each obtained class describes. Furthermore, according to [13], the set of basic classes of any OODN, extended by extracted classes, together with union exploiter, create upper semilattice. Constructed upper semilattice forms a hierarchy of classes, where each class satisfies subsumption relation defined over the hierarchy that makes it possible to find more general class for arbitrary pair of classes. Such approach allows extracting of new knowledge from the basic ones and provides an ability to reconstruct the knowledge base for increasing its compactness.

### III. Universal Exploiters and Knowledge Extraction

As it was shown in [12], some universal exploiters can be efficiently used for KE. According to [12, Th. 1], all possible applications of union exploiter, including all its possible superpositions, to homogeneous classes of objects, which do not have common properties and methods, always generate finite quantity of new classes of objects, which can be precisely calculated.

However, there are situations when homogeneous classes of objects can have common properties and (or) methods. Before we start to consider them, let us make clear what we mean by *type*, *subtype* and *subclass*. As it was mentioned in [13], inhomogeneous class of objects describes at least two different types of objects within one class, where type is defined as follows.

**Definition 1.** Type of objects $t_i$ of arbitrary inhomogeneous class of objects $T = (Core(T), pr_1(T),..., pr_n(T))$, which describes types $t_1,...,t_n$, is a homogeneous class of objects $t_i = (Core(T), pr_i(t_i))$, where $i = \overline{1,n}$.

Consequently, each homogeneous class of objects describes particular type of objects. The definition shows that type and class of objects does not always mean the same, more precisely, homogeneous class of objects is equivalent to type of objects, however inhomogeneous class of objects is not equivalent to type of objects, because it describes some set of types. Now let us define notion of subtype.

**Definition 2.** Arbitrary type of objects $t_1$ is a subtype of arbitrary type of objects $t_2$, i.e. $t_1 \subseteq t_2$ if and only if

$$(P(t_1) \subseteq P(t_2)) \wedge (F(t_1) \subseteq F(t_2)),$$

where $P(t_1)$, $P(t_2)$ are specifications of types $t_1$, $t_2$ and $F(t_1)$, $F(t_2)$ are their signatures.

This definition actually defines the notion of subclass for the case of homogeneous classes, however it is not enough for the inhomogeneous classes of objects. The notion of subclass for inhomogeneous classes was introduced in [13], nevertheless it is restricted and does not take into account some cases, when classes of objects have common properties and methods. Let us consider an example for clear understanding.

Suppose we have three homogeneous classes of objects

$$T_1 = (p_1(T_1),..., p_n(T_1), f_1(T_1),..., f_m(T_1)),$$
$$T_2 = (p_1(T_2),..., p_k(T_2), f_1(T_2),..., f_w(T_2)),$$
$$T_3 = (p_1(T_3),..., p_v(T_3), f_1(T_3),..., f_r(T_3)).$$

Let us assume that

$$p_1(T_1) = p_1(T_2) = p_1(T_3); \; p_2(T_1) = p_2(T_2) = p_2(T_3),$$
$$p_3(T_1) = p_3(T_2) = p_3(T_3); \; p_4(T_1) = p_4(T_2).$$

Using union exploiter, let us compute $T_1 \cup T_2$, $T_1 \cup T_3$ and $T_1 \cup T_2 \cup T_3$, i.e.

$$T_1 \cup T_2 = T_{12} = (Core(T_{12}), pr_1(t_1), pr_2(t_2)) =$$

$$= ((p_1(T_{12}), p_2(T_{12}), p_3(T_{12}), p_4(T_{12})),$$
$$(p_5(t_1),..., p_n(t_1), f_1(t_1),..., f_m(t_1)),$$
$$(p_5(t_2),..., p_k(t_2), f_1(t_2),..., f_w(t_2))).$$
$$T_1 \cup T_3 = T_{13} = (Core(T_{13}), pr_1(t_1), pr_2(t_3)) =$$
$$= ((p_1(T_{13}), p_2(T_{13}), p_3(T_{13})),$$
$$(p_4(t_1),..., p_n(t_1), f_1(t_1),..., f_m(t_1)),$$
$$(p_4(t_3),..., p_v(t_3), f_1(t_3),..., f_r(t_3))).$$
$$T_1 \cup T_2 \cup T_3 = T_{123} =$$
$$= (Core(T_{123}), pr_1(t_1), pr_2(t_2), pr_3(t_3)) =$$
$$= ((p_1(T_{123}), p_2(T_{123}), p_3(T_{123})),$$
$$(p_4(t_1),..., p_n(t_1), f_1(t_1),..., f_m(t_1)),$$
$$(p_4(t_2),..., p_k(t_2), f_1(t_2),..., f_w(t_2)),$$
$$(p_4(t_3),..., p_v(t_3), f_1(t_3),..., f_r(t_3))).$$

According to [13, Def. 12], $T_{13} \subseteq T_{123}$, however $T_{12} \not\subset T_{123}$. Nevertheless, according to Def. 1 and Def. 2,
$$T_1 \subseteq (Core(T_{123}), pr_1(t_1)),$$
$$T_2 \subseteq (Core(T_{123}), pr_1(t_1)).$$

Despite this, [13, Def. 12] is correct for the case when classes of objects have no common properties and methods. Let us assume that classes $T_1$, $T_2$ and $T_3$ do not have common properties and methods, then we have
$$T_1 \cup T_2 = T_{12} = (pr_1(t_1), pr_2(t_2)) =$$
$$= ((p_1(t_1),..., p_n(t_1), f_1(t_1),..., f_m(t_1)),$$
$$(p_1(t_2),..., p_k(t_2), f_1(t_2),..., f_w(t_2))).$$
$$T_1 \cup T_3 = T_{13} = (pr_1(t_1), pr_2(t_3)) =$$
$$= ((p_1(t_1),..., p_n(t_1), f_1(t_1),..., f_m(t_1)),$$
$$(p_1(t_3),..., p_v(t_3), f_1(t_3),..., f_r(t_3))).$$
$$T_1 \cup T_2 \cup T_3 = T_{123} =$$
$$= (pr_1(t_1), pr_2(t_2), pr_3(t_3)) =$$
$$= ((p_1(t_1),..., p_n(t_1), f_1(t_1),..., f_m(t_1)),$$
$$(p_1(t_2),..., p_k(t_2), f_1(t_2),..., f_w(t_2)),$$
$$(p_1(t_3),..., p_v(t_3), f_1(t_3),..., f_r(t_3))).$$

According to [13, Def. 12], $T_{13} \subseteq T_{123}$ and $T_{12} \subseteq T_{123}$, therefore all results, which were presented in [13] are correct. That is why, let us extend the notion of subclass given in [13], using Def. 1 and Def. 2.

**Definition 3.** Arbitrary class of objects $T_1$, which describes types $t_1^1,...,t_n^1$, is a subclass of arbitrary class of objects $T_2$, which describes types $t_1^2,...,t_m^2$, i.e. $T_1 \subseteq T_2$ if and only if $\forall t_i^1 \exists t_j^2 \mid t_i^1 \subseteq t_j^2$, where $i = \overline{1,n}$, $j = \overline{1,m}$ and $n, m \geq 1$.

Now, using this definition for classes $T_{12}$, $T_{13}$ and $T_{123}$ from Example 1, we can conclude that $T_{12} \subseteq T_{123}$ and $T_{13} \subseteq T_{123}$ for both cases, when classes $T_1$, $T_2$ and $T_3$ have common properties and methods and when they do not have them.

Let us consider homogeneous classes of objects $T_1,...,T_n$, which describes types of objects $t_1,...,t_n$. Let us assume that there is such type $t$, that $t \subseteq t_1, t \subseteq t_2,..., t \subseteq t_n$. It means that classes of objects $T_1,...,T_n$ have some common properties and (or) methods. It is clearly, that the application of union exploiter to them will produce a set of new classes of objects. Using this idea, let us formulate and prove the following theorem.

**Theorem 1.** *For any*
$$OODN = (O, C = \{T_1,...,T_n\}, R, E = \{\cup\}, M),$$
*where $T_1,...,T_n$ are homogeneous classes, which describe types of objects $t_1,...,t_n$ and there is a type $t$, such that $t \subseteq t_1, t \subseteq t_2,..., t \subseteq t_n$, all possible applications of union exploiter, including all possible its superpositions, to classes of objects from the set $C$ and obtaining classes of objects using union exploiter, always generate finite quantity of new classes of objects, which can be precisely calculated by the following formula:*
$$q(C_E) = 2^n - n - 1,$$
*where $n = |C|$.*

*Proof:* According to definition of union exploiter for classes of objects [13, Def. 14], the result of union of two arbitrary nonequivalent classes of objects $T_1$ and $T_2$, which describe type of objects $t_1$ and $t_2$ respectively, is inhomogeneous class of objects $T$, which describes both these types. If there is a type $t$, such that $t \subseteq t_1, t \subseteq t_2, ..., t \subseteq t_n$, then class $T$ will have the following structure

$$T = (Core(T), pr_1(t_1), pr_2(t_2)).$$

According to proof of [13, Th. 1], the number of all possible unique classes of objects created from the basic set of classes $C = \{T_1, ..., T_n\}$ using union exploiter can be represented as a combination of $k = \overline{2, n}$ different classes from the set $C$. It is known that

$$\sum_{n=0}^{k} C_n^k = 2^n.$$

However, we cannot create classes of objects, which describe $1$ and $0$ different types, applying union exploiter to the classes of objects from the set $C$, i.e. we do not count $C_n^0$ and $C_n^1$. Therefore, we can conclude that

$$q(C_E) = \sum_{k=0}^{n} C_n^k - C_n^1 - C_n^0 = \sum_{k=2}^{n} C_n^k = 2^n - n - 1. \quad \blacksquare$$

Similarly to [13, Th. 2], we can formulate and prove the following theorem.

**Theorem 2.** *Set of classes of objects*

$$C = \{T_1, ..., T_n, T_{n+1}, ..., T_{2^n-1}\}$$

*of any OODN, extended according to Th. 1, with union exploiter create the join-semilattice $JSL = (C, E = \{\cup, 1\})$, where class $T_{JSL} = T_1 \cup ... \cup T_n$ is its greatest upper bound, i.e. $1$.*

*Proof:* According to the definition of join-semilattice given in [14], [15], it is a system $JSL = (A, \Omega = \{\vee, 1\})$, where $A$ is a poset, $\vee$ is a binary, idempotent, commutative and associative operation and $1$ is an unary operation, which are defined over the set $A$. In addition $\forall a \in A$, $1$ satisfies

$$(L_1): a \vee 1 = 1 \text{ (identity law)}.$$

According to the theorem, carrier of join-semilattice is the set of classes $C$, set of exploiters $E$ contains binary operation $\cup$ and unary operation $1$, which are defined over the set $C$. Therefore $JSL = (C, E = \{\cup, 1\})$, where

$$C = \{T_1, ..., T_n, T_{n+1}, ..., T_{2^n-1}\}.$$

From the [13, Def. 14] it follows, that mentioned properties of $\vee$ are also true for $\cup$, i.e.

1. $T_1 \cup T_1 = T_1$,
2. $T_1 \cup T_2 = T_2 \cup T_1$,
3. $T_1 \cup (T_2 \cup T_3) = (T_1 \cup T_2) \cup T_3$,

where $T_1, T_2, T_3 \in C$. From the definition of $\cup$ it follows that $T_1 \cup T_{JSL} = T_{JSL}$, where $T_{JSL} = T_1 \cup ... \cup T_n$.

Now we need to prove that $C$ is a poset. For this we should define $\forall T_1, T_2 \in C | T_1 \subseteq T_2 \Leftrightarrow T_1 \cup T_2 = T_2$ and show that $\subseteq$ is a relation of partial order under the set $C$. Taking into account that we have two types of classes, we need to define three kinds of $\subseteq$ relation, i.e.

1. homogeneous $\subseteq$ homogeneous,
2. homogeneous $\subseteq$ inhomogeneous,
3. inhomogeneous $\subseteq$ inhomogeneous.

It was done in Def. 2 and Def. 3. Now let us prove reflexivity, anti-symmetry and transitivity of these relations.

1. *Reflexivity:* $T_1 \subseteq T_1 \Leftrightarrow T_1 \cup T_1 = T_1$ follows from idempotency of $\cup$;

2. *Anti-symmetry:*

$$T_1 \subseteq T_2 \Leftrightarrow T_1 \cup T_2 = T_2,$$
$$T_2 \subseteq T_1 \Leftrightarrow T_2 \cup T_1 = T_1$$

and from commutativity of $\cup$, we can conclude that $T_1 = T_2$;

3. *Transitivity:*

$$T_1 \subseteq T_2 \Leftrightarrow T_1 \cup T_2 = T_2, \ T_2 \subseteq T_3 \Leftrightarrow T_2 \cup T_3 =$$
$$= T_3 \Rightarrow (T_1 \cup T_2) \cup T_3 = T_1 \cup (T_2 \cup T_3) =$$
$$= T_1 \cup T_3 = T_3 \Rightarrow T_1 \cup T_3 = T_3 \Leftrightarrow T_1 \subseteq T_3.$$

Therefore,

$$JSL = (C = \{T_1, ..., T_n, T_{n+1}, ..., T_{2^n-1}\}, E = \{\cup, 1\})$$

is a join-semilattice, where class $T_{JSL} = T_1 \cup ... \cup T_n$ is its greatest upper bound, i.e. $1$. $\blacksquare$

Now let us define intersection exploiter for classes of objects, using Def. 1.

**Definition 4.** Intersection of two arbitrary nonequivalent classes of objects $T_1 \cap T_2$, which describe types of objects $t_1^1, ..., t_n^1$ and $t_1^2, ..., t_m^2$, respectively, where $n, m \geq 1$, is inhomogeneous class of objects $T$, which describes types of objects $t_1^3, ..., t_w^3$, where $w \geq 1$, such that

$$(\forall t_k^3, \exists t_i^1 \wedge \exists t_j^2 \mid (t_k^3 \subseteq t_i^1) \wedge (t_k^3 \subseteq t_j^2)) \wedge (\neg \exists t_l' \mid (t_k^3 \subset t_l') \wedge$$
$$\wedge (t_l' \subseteq t_i^1) \wedge (t_l' \subseteq t_j^2)),$$

where $k = \overline{1, w}$, $i = \overline{1, n}$, $j = \overline{1, m}$.

Using this definition, let us formulate and prove the following theorem.

**Theorem 3.** *For any*

$$OODN = (O, C = \{T_1, ..., T_n\}, R, E = \{\cap\}, M),$$

*where $T_1, ..., T_n$ are homogeneous classes, which describe types of objects $t_1, ..., t_n$ and there is a type $t$, such that*

$$((t \subseteq t_1) \wedge ... \wedge (t \subseteq t_n)) \wedge (\neg \exists t' \mid (t \subset t') \wedge$$
$$\wedge (t' \subseteq t_1) \wedge ... \wedge (t' \subseteq t_n)),$$

*all possible applications of intersection exploiter, including all possible its superpositions, to classes of objects from the set $C$ and obtaining classes of objects, using intersection exploiter, always generate finite quantity of new classes of objects, which can be precisely calculated by the following formula:*

$$q(C_E) = 2^n - n - 1,$$

*where $n = |C|$.*

*Proof:* According to Def. 4, the result of intersection of two arbitrary nonequivalent classes of objects $T_1$ and $T_2$ is inhomogeneous class of objects $T$ that describes subtypes, which are common for all types of class $T_1$ and $T_2$ simultaneously.

It is known that the number of all possible unique classes of objects created from the basic set of classes $C = \{T_1, ..., T_n\}$ using intersection exploiter can be represented as a combination of $k = \overline{2, n}$ different classes from the set $C$. It is known that

$$\sum_{n=0}^{k} C_n^k = 2^n.$$

However, intersection exploiter is a binary operation, that is why we cannot count $C_n^0$ and $C_n^1$, therefore

$$q(C_E) = \sum_{k=0}^{n} C_n^k - C_n^1 - C_n^0 = \sum_{k=2}^{n} C_n^k = 2^n - n - 1. \blacksquare$$

Similarly to [13, Th. 2] and Th. 2, we can formulate and prove the following theorem.

**Theorem 4.** *Set of classes of objects*

$$C = \{T_1, ..., T_n, T_{n+1}, ..., T_{2^n - 1}\}$$

*of any OODN, extended according to Th. 3, with exploiter $\cap$ create the meet-semilattice $MSL = (C, E = \{\cap, 0\})$, where class $T_{MSL} = T_1 \cap ... \cap T_n$ is its least lower bound, i.e. $0$.*

*Proof:* According to definition of meet-semilattice given in [14], [15], it is a system $MSL = (A, \Omega = \{\wedge, 0\})$, where $A$ is a poset, $\wedge$ is binary, idempotent, commutative and associative operation and $0$ is unary operation, which are defined over the set $A$. In addition, $\forall a \in A$, $0$ satisfies

$$(L_1): a \wedge 0 = 0 \text{ (identity law)}.$$

According to the theorem, carrier of meet-semilattice is the set of classes $C$, set of exploiters $E$ contains binary operation $\cap$ and unary operation $0$, which are defined over the set $C$. Therefore, $MSL = (C, E = \{\cap, 0\})$, where

$$C = \{T_1, ..., T_n, T_{n+1}, ..., T_{2^n - 1}\}.$$

From the Def. 4 it follows, that all mentioned properties of $\wedge$ are also true for $\cap$, i.e.

1. $T_1 \cap T_1 = T_1$,
2. $T_1 \cap T_2 = T_2 \cap T_1$,
3. $T_1 \cap (T_2 \cap T_3) = (T_1 \cap T_2) \cap T_3$,

where $T_1, T_2, T_3 \in C$. From the definition of $\cap$ it follows that, $T_1 \cap T_{MSL} = T_{MSL}$, where $T_{MSL} = T_1 \cap ... \cap T_n$.

Now we need to prove that $C$ is a poset. For this we should define $\forall T_1, T_2 \in C \mid T_1 \subseteq T_2 \Leftrightarrow T_1 \cap T_2 = T_1$ and show that $\subseteq$ is a relation of partial order under the set $C$. Taking into account that we have two types of classes, we need to define three kinds of $\subseteq$ relation, i.e.

1. homogeneous $\subseteq$ homogeneous,
2. homogeneous $\subseteq$ inhomogeneous,
3. inhomogeneous $\subseteq$ inhomogeneous.

It was done in Def. 2 and Def. 3. Now let us prove reflexivity, anti-symmetry and transitivity of these relations.

1. *Reflexivity:* $T_1 \subseteq T_1 \Leftrightarrow T_1 \cap T_1 = T_1$ follows from idempotency of $\cap$;
2. *Anti-symmetry:* $T_1 \subseteq T_2 \Leftrightarrow T_1 \cap T_2 = T_1$, and $T_2 \subseteq T_1 \Leftrightarrow T_2 \cap T_1 = T_2$ and from commutativity of $\cap$, we can conclude that $T_1 = T_2$;

3. *Transitivity:*

$$T_2 \subseteq T_1 \Leftrightarrow T_1 \cap T_2 = T_2, \ T_3 \subseteq T_2 \Leftrightarrow T_2 \cap T_3 =$$
$$= T_3 \Rightarrow (T_1 \cap T_2) \cap T_3 = T_1 \cap (T_2 \cap T_3) =$$
$$= T_1 \cap T_3 = T_3 \Rightarrow T_1 \cap T_3 = T_3 \Leftrightarrow T_3 \subseteq T_1.$$

Therefore,

$$MSL = (C = \{T_1, ..., T_n, T_{n+1}, ..., T_{2^n-1}\}, E = \{\cap, 0\})$$

is a meet-semilattice, where class $T_{MSL} = T_1 \cap ... \cap T_n$ is its least lower bound, i.e. $0$. ∎

Using Th. 1 and Th. 3, let us formulate and prove the following theorem.

**Theorem 5.** *For any $OODN = (O, C, R, E, M)$, where $C = \{T_1, ..., T_n, T_{n+1}, ..., T_{2^n-1}\}$, $E = \{\cup, \cap\}$, and $T_1, ..., T_n$ are homogeneous classes, which describe types of objects $t_1, ..., t_n$ and there is a type $t$, such that*

$$((t \subseteq t_1) \wedge ... \wedge (t \subseteq t_n)) \wedge (\neg \exists t' \mid (t \subset t') \wedge$$
$$\wedge (t' \subseteq t_1) \wedge ... \wedge (t' \subseteq t_n)),$$

*all possible applications of union and intersection exploiters, including all possible their superpositions, to classes of objects from the set $C$ and obtaining classes of objects, using these exploiters respectively, always generate finite quantity of new classes of objects, which can be precisely calculated by the following formula:*

$$q(C_E) = 2^{n+1} - 2(n+1),$$

*where $n = |C|$.*

*Proof:* Proof of the theorem follows from proofs of Th. 1 and Th. 3, i.e.

$$q(C_E) = 2 \sum_{k=0}^{n} C_n^k - C_n^1 - C_n^0 = 2 \sum_{k=2}^{n} C_n^k = 2^{n+1} - 2(n+1),$$

where $n = |C|$. ∎

Similarly to Th. 2 and Th. 4, we can formulate and prove the following important theorem.

**Theorem 6.** *Set of classes of objects*

$$C = \{T_1, ..., T_n, T_{n+1}, ..., T_{2^n-1}, T_{2^n}, ..., T_{2^{n+1}-2(n+1)}\}$$

*of any OODN, extended according to Th. 5, with exploiters $\cup$, $\cap$ create the complete lattice $L = (C, E = \{\cup, \cap, 1, 0\})$, where class $T_{JSM} = T_1 \cup ... \cup T_n$ is the greatest upper bound,*

i.e. $1$ and class $T_{MSL} = T_1 \cap ... \cap T_n$ is its least lower bound, i.e. $0$.

*Proof:* According to definition of complete lattice given in [14], [15], it is a system $L = (A, \Omega = \{\vee, \wedge, 1, 0\})$, where $A$ is a poset and $\vee$, $\wedge$, $1$ and $0$ satisfy, for all $a, b, c \in A$:

$(L_1): (a \vee b) \vee c = a \vee (b \vee c)$ (associative laws)

$(a \wedge b) \wedge c = a \wedge (b \wedge c)$

$(L_2): a \vee b = b \vee a$ (commutative laws)

$a \wedge b = b \wedge a$

$(L_3): a \vee a = a$ (idempotency laws)

$a \wedge a = a$

$(L_4): a \vee (a \wedge b) = a$ (absorption laws)

$a \wedge (a \vee b) = a$

$(L_5): a \vee 0 = a$ (identity laws)

$a \wedge 1 = a$

$a \wedge 0 = 0$

$a \vee 1 = 1$

According to the theorem, carrier of the lattice is the set of classes $C$, set of exploiters $E$ contains two binary operations $\cup$, $\cap$ and two unary operations $1$ and $0$, which are defined over the set $C$. Therefore, $L = (C, E = \{\cup, \cap, 1, 0\})$, where

$$C = \{T_1, ..., T_n, T_{n+1}, ..., T_{2^n-1}, T_{2^n}, ..., T_{2^{n+1}-2(n+1)}\}$$

Facts that

1. $(C, \subseteq)$ is a poset,
2. $\cup$ and $\cap$ satisfy the laws $(L_1) - (L_3)$,
3. $T_{JSL} = T_1 \cup ... \cup T_n$ is $1$ of join-semilattice,
4. $T_{MSL} = T_1 \cap ... \cap T_n$ is $0$ of meet-semilattice,

were shown in the proves of Th. 2 and Th. 4.

From the [13, Def. 14] and Def. 4 it follows, that $T_1 \cup T_{MSL} = T_1$, $T_1 \cap T_{JSL} = T_1$, $T_1 \cap T_{MSL} = T_{MSL}$, and $T_1 \cup T_{JSL} = T_{JSL}$, where $T_1 \in C$, $T_{JSL} = T_1 \cup ... \cup T_n$, $T_{MSL} = T_1 \cap ... \cap T_n$. Therefore, $L = (C, E = \{\cup, \cap, 1, 0\})$ is a complete lattice, where

$$C = \{T_1,...,T_n, T_{n+1},...,T_{2^n-1}, T_{2^n},...,T_{2^{n+1}-2(n+1)}\},$$

and $T_{JSL} = T_1 \cup ... \cup T_n$ is its greatest upper bound, i.e. 1 and $T_{MSL} = T_1 \cap ... \cap T_n$ is its least lower bound, i.e. 0. ∎

## IV. EXPLOITERS-BASED KNOWLEDGE EXTRACTION

Let us consider classes of objects, which describe such types of convex polygons as square $(S)$, rhombus $(Rb)$, parallelogram $(P)$, and rectangle $(Rt)$. Let us define for them an OODN

$$Quadrangle = (O, C, R, E, M).$$

For this purpose, we need to define set of classes of objects $C = \{S, Rb, P, Rt\}$ and set of exploiters $E = \{\cup, \cap\}$. Sets $O$ and $R$ will be undefined, because of the lack of information. In addition, we do not define the set of modifiers $M$, because it is not necessary within consideration of exploiters-based KE. Suppose classes from the set $C$ have following structures

$$S = (p_1(S) = (4, sides),$$

$$p_2(S) = (4, angles)$$

$$p_3(S) = ((v_1(p_3(S)), cm), (v_2(p_3(S)), cm),$$

$$(v_3(p_3(S)), cm), (v_4(p_3(S)), cm)),$$

$$p_4(S) = (90°, 90°, 90°, 90°),$$

$$p_5(S) = vf_5(S) = 1,$$

$$p_6(S) = vf_6(S) = 1,$$

$$p_7(S) = vf_7(S) = 1,$$

$$f_1(S) = (v_1(p_3(S)) \cdot 4, cm),$$

$$f_2(S) = (v_1(p_3(S))^2, cm^2)),$$

where $p_1(S)$ – quantity of sides, $p_2(S)$ – quantity of angles, $p_3(S)$ – sizes of sides, $p_4(S)$ – measures of internal angles, $p_5(S)$ – verification function, which defines property «sum of internal angles is equal to $360°$», i.e.

$$vf_5(S) : p_5(S) \to \{0,1\},$$

where

$$p_5(S) = (v_1(p_4(S)) + v_2(p_4(S)) + v_3(p_4(S)) +$$

$$+ v_4(p_4(S)) = 360),$$

$p_6(S)$ – verification function, which defines property «all sides are equal», i.e. $vf_6(S) : p_6(S) \to \{0,1\}$, where

$$p_6(S) = (v_1(p_3(S)) = v_2(p_3(S)) = v_3(p_3(S)) =$$

$$= v_4(p_3(S))),$$

$p_7(S)$ – verification function, which defines property «all angles are equal to $90°$», i.e. $vf_7(S) : p_7(S) \to \{0,1\}$, where

$$p_7(S) = (v_1(p_4(S)) = v_2(p_4(S)) = v_3(p_4(S)) =$$

$$= v_4(p_4(S)) = 90),$$

$f_1(S)$ – method for perimeter computing, and $f_2(S)$ – method for area computing;

$$Rb = (p_1(Rb) = (4, sides),$$

$$p_2(Rb) = (4, angles),$$

$$p_3(Rb) = ((v_1(p_3(Rb)), cm), (v_2(p_3(Rb)), cm),$$

$$(v_3(p_3(Rb)), cm), (v_4(p_3(Rb)), cm)),$$

$$p_4(Rb) = ((v_1(p_4(Rb)), °), (v_2(p_4(Rb)), °),$$

$$(v_3(p_4(Rb)), °), (v_4(p_4(Rb)), °)),$$

$$p_5(Rb) = vf_5(Rb) = 1,$$

$$p_6(Rb) = vf_6(Rb) = 1,$$

$$f_1(Rb) = (v_1(p_3(Rb)) \cdot 4, cm),$$

$$f_2(Rb) = (v_1(p_3(Rb))^2 \cdot \sin(v_1(p_4(Rb))), cm^2)),$$

where $p_1(Rb)$ – quantity of sides, $p_2(Rb)$ – quantity of angles, $p_3(Rb)$ – sizes of sides, $p_4(Rb)$ – measures of internal angles, $p_5(Rb)$ – verification function, which defines property «sum of internal angles is equal to $360°$», i.e.

$$vf_5(Rb) : p_5(Rb) \to \{0,1\},$$

where

$$p_5(Rb) = (v_1(p_4(Rb)) + v_2(p_4(Rb)) + v_3(p_4(Rb)) +$$

$$+ v_4(p_4(Rb)) = 360),$$

$p_6(Rb)$ – verification function, which defines property «all sides are equal», i.e. $vf_6(Rb): p_6(Rb) \to \{0,1\}$, where

$$p_6(Rb) = (v_1(p_3(Rb)) = v_2(p_3(Rb)) = v_3(p_3(Rb)) =$$
$$= v_4(p_3(Rb))),$$

$f_1(Rb)$ – method for perimeter computing, and $f_2(Rb)$ – method for area computing;

$P = (p_1(P) = (4, sides),$

$\quad p_2(P) = (4, angles),$

$\quad p_3(P) = ((v_1(p_3(P)), cm), (v_2(p_3(P)), cm),$
$\quad\quad (v_3(p_3(R)), cm), (v_4(p_3(R)), cm)),$

$\quad p_4(P) = ((v_1(p_4(P)),°), (v_2(p_4(P)),°),$
$\quad\quad (v_3(p_4(P)),°), (v_4(p_4(P)),°)),$

$\quad p_5(P) = vf_5(P) = 1,$

$\quad p_6(P) = vf_6(P) = 1,$

$\quad p_7(P) = vf_7(P) = 1,$

$\quad f_1(P) = (2 \cdot (v_1(p_3(P)) + v_2(p_3(P))), cm),$

$\quad f_2(P) = (v_1(p_3(P)) \cdot v_2(p_3(P)) \cdot$
$\quad\quad \cdot \sin(v_1(p_4(P))), cm^2)),$

where $p_1(P)$ – quantity of sides, $p_2(P)$ – quantity of angles, $p_3(P)$ – sizes of sides, $p_4(P)$ – measures of internal angles, $p_5(P)$ – verification function, which defines property «sum of internal angles is equal to $360°$», i.e.

$$vf_5(P): p_5(P) \to \{0,1\},$$

where

$$p_5(P) = (v_1(p_4(P)) + v_2(p_4(P)) + v_3(p_4(P)) +$$
$$+ v_4(p_4(P)) = 360),$$

$p_6(P)$ – verification function, which defines property «opposite sides are parallel», i.e. $vf_6(P): p_6(P) \to \{0,1\}$, where

$$p_6(P) = (v_1(p_4(P)) = v_3(p_4(P))) \wedge (v_2(p_4(P)) =$$
$$= v_4(p_4(P))),$$

$p_7(P)$ – verification function, which defines property «opposite sides are equal», i.e. $vf_7(P): p_7(P) \to \{0,1\}$, where

$$p_7(P) = (v_1(p_3(P)) = v_3(p_3(P))) \wedge (v_2(p_3(P)) =$$
$$= v_4(p_3(P))),$$

$f_1(P)$ – method for perimeter computing, and $f_2(P)$ – method for area computing;

$Rt = (p_1(Rt) = (4, sides),$

$\quad p_2(Rt) = (4, angles),$

$\quad p_3(Rt) = ((v_1(p_3(Rt)), cm), (v_2(p_3(Rt)), cm),$
$\quad\quad (v_3(p_3(Rt)), cm), (v_3(p_3(Rt)), cm)),$

$\quad p_4(Rt) = ((90,°), (90,°), (90,°), (90,°)),$

$\quad p_5(Rt) = vf_5(Rt) = 1,$

$\quad p_6(Rt) = vf_6(Rt) = 1,$

$\quad f_1(Rt) = (2 \cdot (v_1(p_3(Rt)) + v_2(p_3(Rt))), cm),$

$\quad f_2(Rt) = (v_1(p_3(Rt)) \cdot v_2(p_3(Rt)), cm^2)),$

where $p_1(Rt)$ – quantity of sides, $p_2(Rt)$ – quantity of angles, $p_3(Rt)$ – sizes of sides, $p_4(Rt)$ – measures of internal angles, $p_5(Rt)$ – verification function, which defines property «sum of internal angles is equal to $360°$», i.e.

$$vf_5(Rt): p_5(Rt) \to \{0,1\},$$

where

$$p_5(Rt) = (v_1(p_4(Rt)) + v_2(p_4(Rt)) + v_3(p_4(Rt)) +$$
$$+ v_4(p_4(Rt)) = 360),$$

$p_6(Rt)$ – verification function, which defines property «opposite sides are equal», i.e. $vf_6(Rt): p_6(Rt) \to \{0,1\}$, where

$$p_6(Rt) = (v_1(p_3(Rt)) = v_3(p_3(Rt))) \wedge (v_2(p_3(Rt)) =$$
$$= v_4(p_3(Rt))),$$

$f_1(Rt)$ – method for perimeter computing, and $f_2(Rt)$ – method for area computing.

We have defined OODN for early mentioned types of convex polygons. It is clear, that all elements of the set $C$ represent basic knowledge. Let us apply union and intersection exploiters to them and obtain all possible new classes of objects. According to [13, Def. 14],

$$S \cup Rb = SRb_\cup = (Core(SRb_\cup), pr_1(S), pr_2(Rb)),$$

where

$$Core(SRb_\cup) = (p_1(SRb_\cup), p_2(SRb_\cup), p_3(SRb_\cup),$$
$$p_4(SRb_\cup), f_1(SRb_\cup)),$$

where $p_1(SRb_\cup)$ – quantity of sides, $p_2(SRb_\cup)$ – quantity of angles, $p_3(SRb_\cup)$ – verification function, which defines property «sum of internal angles is equal to $360°$», i.e.

$$vf_3(SRb_\cup) : p_3(SRb_\cup) \to \{0,1\},$$

where

$$p_3(SRb_\cup) = (v_1(p_4(t_i)) + v_2(p_4(t_i)) + v_3(p_4(t_i)) +$$
$$+ v_4(p_4(t_i)) = 360), \ i \in \{S, Rb\},$$

$p_4(SRb_\cup)$ – verification function, which defines property «all sides are equal», i.e. $vf_4(SRb_\cup) : p_4(SRb_\cup) \to \{0,1\}$, where

$$p_4(SRb_\cup) = (v_1(p_3(t_i)) = v_2(p_3(t_i)) = v_3(p_3(t_i)) =$$
$$= v_4(p_3(t_i))), \ i \in \{S, Rb\},$$

$f_1(SRb_\cup)$ – method for perimeter computing, which is defined as follows $f_1(SRb_\cup) = (4 \cdot v_1(p_3(t_i)), cm)$, where $i \in \{S, Rb\}$.

Projections $pr_1(S)$ and $pr_2(Rb)$ have the following structure

$$pr_1(S) = (p_5(S), p_6(S), p_7(S), f_2(S)),$$
$$pr_2(Rb) = (p_5(Rb), p_6(Rb), f_2(Rb)),$$

where $p_5(S)$ – sizes of sides, $p_6(S)$ – measures of internal angles, $p_7(S)$ – verification function, which defines property «all angles are equal to $90°$», $f_2(S)$ – method for area computing, $p_5(Rb)$ – sizes of sides, $p_6(Rb)$ – measures of internal angles, $f_2(Rb)$ – method for area computing.

Structure of the $Core(SRb_\cup)$ follows from the following equalities

$$p_1(S) = p_1(Rb), \ p_2(S) = p_2(Rb), \ p_5(S) = p_5(Rb),$$
$$p_6(S) = p_6(Rb), \ f_1(S) = f_1(Rb).$$

Indeed, according to [13, Def 4], $p_1(S) = p_1(Rb)$ and $p_2(S) = p_2(Rb)$, i.e.

$$(v_1(p_1(S)), u_1(p_1(S))) = (v_1(p_1(Rb)), u_1(p_1(Rb))) =$$
$$= (4, sides),$$
$$(v_1(p_2(S)), u_1(p_2(S))) = (v_1(p_2(Rb)), u_1(p_2(Rb))) =$$
$$= (4, angles).$$

Form the [13, Def. 5] it follows that $p_5(S) = p_5(Rb)$ and $p_6(S) = p_6(Rb)$, i.e.

$$\left( vf_5^S(S) = vf_5^{Rb}(S) \right) \wedge \left( vf_5^S(Rb) = vf_5^{Rb}(Rb) \right),$$

that can be computed in the following way

$$vf_5^S(S) = v_1(p_4(S)) + v_2(p_4(S)) + v_3(p_4(S)) +$$
$$+ v_4(p_4(S)) = 360,$$
$$vf_5^{Rb}(S) = v_1(p_4(S)) + v_2(p_4(S)) + v_3(p_4(S)) +$$
$$+ v_4(p_4(S)) = 360,$$
$$vf_5^S(Rb) = v_1(p_4(Rb)) + v_2(p_4(Rb)) + v_3(p_4(Rb)) +$$
$$+ v_4(p_4(Rb)) = 360,$$
$$vf_5^{Rb}(Rb) = v_1(p_4(Rb)) + v_2(p_4(Rb)) + v_3(p_4(Rb)) +$$
$$+ v_4(p_4(Rb)) = 360;$$
$$\left( vf_6^S(S) = vf_6^{Rb}(S) \right) \wedge \left( vf_6^S(Rb) = vf_6^{Rb}(Rb) \right)$$
$$vf_6^S(S) = (v_1(p_3(S)) = v_2(p_3(S)) = v_3(p_3(S)) =$$
$$= v_4(p_3(S))),$$
$$vf_6^{Rb}(S) = (v_1(p_3(S)) = v_2(p_3(S)) = v_3(p_3(S)) =$$
$$= v_4(p_3(S))),$$
$$vf_6^S(Rb) = (v_1(p_3(Rb)) = v_2(p_3(Rb)) = v_3(p_3(Rb)) =$$
$$= v_4(p_3(Rb))),$$

$$vf_6^{Rb}(Rb) = (v_1(p_3(Rb)) = v_2(p_3(Rb)) = v_3(p_3(Rb)) =$$
$$= v_4(p_3(Rb))).$$

As the result, in both cases we have $(1=1) \wedge (1=1)$, i.e. $1 \wedge 1 = 1$.

From the [13, Def. 7] it follows that $f_1(S) = f_1(Rb)$, i.e.
$$\left(f_1^S(S) = f_1^{Rb}(S)\right) \wedge \left(f_1^S(Rb) = f_1^{Rb}(Rb)\right),$$

that can be calculated in the following way
$$f_1^S(S) = (v_1(p_3(S)) \cdot 4, cm),$$
$$f_1^{Rb}(S) = (v_1(p_3(S)) \cdot 4, cm),$$
$$f_1^S(Rb) = (v_1(p_3(Rb)) \cdot 4, cm),$$
$$f_1^{Rb}(Rb) = (v_1(p_3(Rb)) \cdot 4, cm),$$

as the result we have
$$\left((v_1(p_3(S)) \cdot 4, cm) = (v_1(p_3(S)) \cdot 4, cm)\right) \wedge$$
$$\wedge \left((v_1(p_3(Rb)) \cdot 4, cm) = (v_1(p_3(Rb)) \cdot 4, cm)\right),$$

i.e. $1 \wedge 1 = 1$.

According to [13, Def. 14], the class of objects $SRb_\cup$ is the result of application of union exploiter to classes of objects $S$ and $Rb$. From the Def. 4, we can conclude, that the result of application of intersection exploiter to these classes is equal to the core of their union, i.e.
$$S \cap Rb = SRb_\cap = Core(SRb_\cup).$$

In the result of all possible applications of union and intersection exploiters we obtained such 6 classes, that each class describes 2 different types of objects $SRb_\cup$, $SP_\cup$, $SRt_\cup$, $RbP_\cup$, $RbRt_\cup$, $PRt_\cup$ such 4 classes, that each class describes 3 different types of objects $SRbP_\cup$, $SRbRt_\cup$, $SPRt_\cup$, $RbPRt_\cup$ and 1 class, that describes 4 different types of objects $SRbPRt_\cup$. In addition, we obtained such 6 classes, that each class describes intersection of 2 different types of objects $SRb_\cap$, $SP_\cap$, $SRt_\cap$, $RbP_\cap$, $RbRt_\cap$, $PRt_\cap$, such 4 classes, that each class describes intersection of 3 different types of objects $SRbP_\cap$, $SRbRt_\cap$, $SPRt_\cap$, $RbPRt_\cap$, and 1 class, that describes intersection f 4 different types of objects $SRbPRt_\cap$.

Using exploiters $\cup$ and $\cap$, we have extended the set $C$ by adding 22 new classes of objects, i.e.

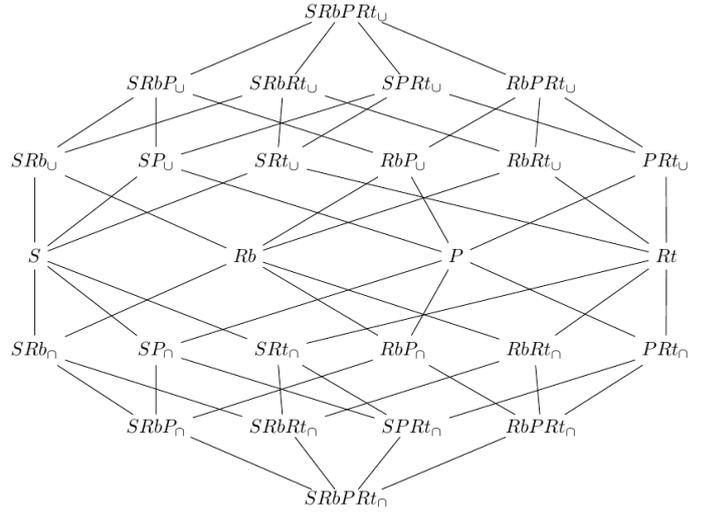

Fig. 1. Complete lattice created by the set of classes and set of exploiters.

$$C = \{S, Rb, P, Rt, SRb_\cup, ..., PRt_\cup, SRbP_\cup, ..., RbPRt_\cup,$$
$$SRbPRt_\cup, SRb_\cap, ..., PRt_\cap, ..., SRbP_\cap, ..., RbPRt_\cap,$$
$$SRbPRt_\cap\}.$$

According to Th. 6, the set $C$ together with exploiters $\cup$ and $\cap$ create the complete lattice $L = (C, E = \{\cup, \cap, 1, 0\})$, where $SRbPRt_\cup$ is its greatest upper bound, i.e. $1$ and $SRbPRt_\cap$ is its least lower bound, i.e. $0$. This lattice can be graphically represented as it is shown on Fig. 1.

In addition, we define the set of relations $R$, by adding 96 new relations, namely 56 relations for classes $S, ..., Rt$, 32 for classes $SRb_\cup, ..., PRt_\cup$ and $SRb_\cap, ..., PRt_\cap$, 8 for classes $SRbP_\cup, ..., RbPRt_\cup$ and $SRbP_\cap, ..., RbPRt_\cap$.

Analyzing Fig. 1, we can see that obtained lattice defines hierarchy of classes with determined subsumption relation $\subseteq$. It allows performing of subsumption reasoning for information classifying and retrieving. Moreover, obtained hierarchy is protected from ambiguity problem, because all classes, except basic ones, are inhomogeneous.

Join-semilattice of the lattice $L$ contains inhomogeneous classes of objects, which define all possible sets of objects of different types, which can be obtained from the basic classes of objects $S$, $Rb$, $P$ and $Rt$. Meet-semilattice of the lattice $L$ contains inhomogeneous classes of objects, which define common subtypes for basic classes.

The greatest upper bound $SRbPRt_\cup$ of the lattice $L$ gives an opportunity to represent and to store the knowledge in the database in more efficient way by storing only one class $SRbPRt_\cup$ instead of four basic classes of objects. Moreover, such storing requires less memory resources then storing of $S$, $Rb$, $P$ and $Rt$, because instead of storing of $26$ properties and $8$ methods, it is possible to store only $17$ properties and $5$ methods.

We can conclude that during KE using universal exploiters we have obtained $22$ new classes of objects, $96$ new relations among them, defined hierarchy of classes with determined subsumption relation $\subseteq$. Using obtained knowledge it is possible to restore basic knowledge in database more efficiently and perform subsumption reasoning within the constructed hierarchy of classes.

## V. Conclusions and Outlook

Invention of KE techniques is very crucial for future development of KRFs and area of KR in general. In this paper the main attention was paid to consideration and extension of KE method within such object-oriented KRF as object-oriented dynamic networks. The main idea of proposed approach is usage of universal exploiters, which allow generation of new classes of objects and relations among them.

The main achievement of the paper is proof of useful features of union and intersection exploiters, which allow extending set of basic classes and create complete lattice. Proposed approach has the following features:

- ability to calculate before the generation:
  o quantity of new classes, which can be generated, using proposed approach,
  o quantity of different types, which each of obtained classes describes;
- extension of the sets of basic classes and relations by adding new classes of objects and relations among them;
- construction of defined hierarchy of classes with determined subsumption relation $\subseteq$, which allows performing of subsumption reasoning for information classifying and retrieving;
- more efficient restoring of basic knowledge within the database;
- avoidance of inheritance problems, in particular ambiguity problem in the case of multiple inheritance.

However, despite all noted advantages, proposed approach requires further research, at least in the following directions:

- using of useful properties of complete lattices;
- adaptation to different kinds of knowledge sources;
- extension to the case of fuzzy knowledge;
- adaptation and usage in other known object-oriented knowledge representation formalisms.

References

[1] D. S. Touretzky, *The mathematics of inheritance systems*. Los Altos, CA, USA: Morgan Kaufmann Publishers, 1986.

[2] R. Al-Asady, *Inheritance theory: an artificial intelligence approach*. Norwood, NJ, USA: Ablex Publishing Corporation, 1995.

[3] D. Terletskyi, "Inheritance in object-oriented knowledge representation," in *Inform. and Software Techn., Commun. in Comput. and Inform. Sci.*, vol. 538, G. Dregvaite and R. Damasevicius Eds., AG, Switzerland: Springer, 2015, pp. 293–305.

[4] S. Hellmann, J. Unbehauen *et al.*, "Report on knowledge extraction from structured sources," Technical Report LOD2 D3.1.1, 2011

[5] J. Unbehauen, S. Hellmann, S. Auer and C. Stadler, "Knowledge extraction from structured sources," in *Search Computing: Broadening Web Search, Lecture Notes in Comput. Sci.*, vol. 7538, S. Ceri and M. Brambilla Eds. Berlin, Germany: Springer, 2012, pp. 34–52.

[6] N. Takhirov, "Extracting knowledge for cultural heritage knowledge base population," Ph.D. dissertation, Dept. Comput. and Inform. Sci., Norwegian Univ. of Sci. and Technol., Trondheim, Norway, 2013.

[7] J. Polpinij, "Ontology-based knowledge discovery from unstructured and semi-structured text," Ph.D. dissertation, School of Comput. Sci. and Software Eng., Univ. of Wollongong, Wollongong, NSW, Australia, 2014.

[8] Z. Ma, F. Zhang, L. Yan and J. Cheng, *Fuzzy knowledge management for the semantic web*. Berlin, Germany: Springer, 2014.

[9] L. Yan, Z. Ma and F. Zhang, Fuzzy *XML data management*. Berlin, Germany: Springer, 2014.

[10] D. Terletskyi and A. Provotar, "Object-oriented dynamic networks," in *Computational Models for Bus. and Eng. Domains, Int. Book Series Inform. Sci. & Computing*, vol. 30, G. Setlak and K. Markov Eds. Rzeszow, Poland: ITHEA, 2014, pp. 123–136.

[11] D. A. Terletskyi and A. I. Provotar, "Fuzzy object-oriented dynamic networks. I," *Cybern. and Syst. Anal.*, vol. 51, no. 1, pp. 34–40, Jan. 2015.

[12] D. A. Terletskyi and A. I. Provotar, "Fuzzy object-oriented dynamic networks. II," *Cybern. and Syst. Anal.*, vol. 52, no. 1, pp. 38–45, Jan. 2016.

[13] D. Terletskyi, "Exploiters-based knowledge extraction in object-oriented knowledge representation," in Proc. 24th Int. *Workshop Concurrency, Specification & Programming*, Rzeszow, 2015, vol. 2, pp. 211–221.

[14] G. Birkhoff, *Lattice theory*. 3rd revised ed. New York, NY, USA: American Mathematical Society Colloquium Publications, 1967.

[15] B. A. Davey and H. A. Priestley, *Introduction to lattices and order*. 2nd ed. New York, NY, USA: Cambridge University Press, 2002.